\newcommand{\CLASSINPUTtoptextmargin}{57pt}
\newcommand{\CLASSINPUTbottomtextmargin}{43pt}
\documentclass[letterpaper, 10 pt, journal, twoside]{IEEEtran}

\usepackage[utf8]{inputenc}
\usepackage[T1]{fontenc}
\usepackage{pdfx}

\usepackage{amsfonts}
\usepackage{amsmath}
\usepackage{amssymb}
\usepackage{algorithmic}
\usepackage{graphicx}
\usepackage{textcomp}
\usepackage{xcolor}

\usepackage[square,numbers,sort&compress]{natbib}

\usepackage{blindtext}
\usepackage{inconsolata}
\usepackage{hyperref}

\definecolor[named]{xGreen}{HTML}{60B950}
\definecolor[named]{xBlue}{HTML}{18647E}
\definecolor[named]{xOrange}{HTML}{FF9A20}
\definecolor[named]{xGray}{HTML}{808080}
\definecolor[named]{xRed}{HTML}{A30B37}
\definecolor[named]{xDarkBlue}{HTML}{0055C9}

\definecolor[named]{TLDRViolet}{HTML}{800080}
\definecolor[named]{MissingCyan}{HTML}{47D4FF}
\definecolor[named]{TODORed}{HTML}{A30B37}
\definecolor[named]{SkeletonGray}{HTML}{808080}

\definecolor[named]{JennGreen}{HTML}{21D19F}
\definecolor[named]{SiddBlue}{HTML}{18647E}

\definecolor[named]{BlakeMagenta}{HTML}{F26DF9}
\definecolor[named]{DorsaOrange}{HTML}{FF9A20}

\hypersetup{
    colorlinks=true,
    linkcolor=xGreen,
    filecolor=magenta,      
    urlcolor=xDarkBlue,
    citecolor=xGreen,
}

\def\Snospace~{\S{}}

\def \papermode{final}  

\def \draftmode{draft}

\ifx \papermode \draftmode
    
    \newcommand{\needcite}{{\color{MissingCyan}[NEED CITE]}}
    \newcommand{\tldr}[1]{{\color{TLDRViolet}{[TL;DR] :: \textit{#1}}}}

    \newcommand{\skeleton}[1]{{\color{SkeletonGray} #1}}

    \newcommand{\makecomment}[3]{{\color{#2}[\textbf{#1}]: #3}}

\else
    
    \newcommand{\needcite}[1]{}
    \newcommand{\tldr}[1]{}
    \newcommand{\pending}[1]{}
    \newcommand{\skeleton}[1]{}
    \newcommand{\makecomment}[3]{}
\fi

\newcommand\blfootnote[1]{%
  \begingroup
  \renewcommand\thefootnote{}\footnote{#1}%
  \addtocounter{footnote}{-1}%
  \endgroup
}

\newcommand{\algname}{Proactive Voice}
\newcommand{\algabbr}{ProVox}

\begin{document}

\title{\vspace{+3pt} \huge ProVox: Personalization and Proactive Planning \\ for Situated Human-Robot Collaboration 
}


\author{
    \IEEEauthorblockN{Jennifer Grannen\IEEEauthorrefmark{2},
        Siddharth Karamcheti\IEEEauthorrefmark{2},
        Blake Wulfe\IEEEauthorrefmark{3},
        Dorsa Sadigh\IEEEauthorrefmark{2}
        \vspace{-0.5cm}
    }
}
\maketitle

\markboth{IEEE Robotics and Automation Letters. Preprint Version. Accepted May, 2025}
{Grannen \MakeLowercase{\textit{et al.}}: ProVox: Personalization and Proactive Planning for Situated Human-Robot Collaboration}

%

\blfootnote{
Manuscript received: December 12, 2024; Revised: March 31, 2025; Accepted: May 29, 2025. 

This paper was recommended for publication by
Editor Angelika Peer upon evaluation of the Associate Editor and Reviewers’ comments.

This work was supported by the Toyota Research Institute (TRI), the DARPA Friction for Accountability in Conversational Transactions (FACT) Program, the AFOSR Young Investigator Program, the Stanford Institute for Human-Centered AI (HAI), the Cooperative AI Foundation, the NSF (Awards \#1941722, \#2006388, \#2125511), the Office of Naval Research (ONR Award \#N000142112298), and DARPA (Grant \#W911NF2210214). 

The authors are with \IEEEauthorrefmark{2}Stanford University and \IEEEauthorrefmark{3}Toyota Research Institute. 
\texttt{jgrannen@stanford.edu}, \texttt{skaramcheti@cs.stanford.edu}, \texttt{blake.wulfe@tri.global}, and \texttt{dorsa@cs.stanford.edu}. 
}

\begin{abstract}
Collaborative robots must quickly adapt to their partner's intent and preferences to proactively identify helpful actions. This is especially true in situated settings where human partners can continually teach robots new high-level behaviors, visual concepts, and physical skills (e.g., through demonstration), growing the robot's capabilities as the human-robot pair work together to accomplish diverse tasks. In this work, we argue that robots should be able to \textit{infer their partner's goals} from early interactions and use this information to \textit{proactively plan behaviors} ahead of explicit instructions from the user. Building from the strong commonsense priors and steerability of large language models, we introduce \algabbr{} (``\algname{}''), a novel framework that enables robots to efficiently personalize and adapt to individual collaborators. We design a \textit{meta-prompting protocol} that empowers users to communicate their distinct preferences, intent, and expected robot behaviors ahead of starting a physical interaction. \algabbr{} then uses the personalized prompt to condition a \textit{proactive language model task planner} that anticipates a user's intent from the current interaction context and robot capabilities to suggest helpful actions; in doing so, we alleviate user burden, minimizing the amount of time partners spend explicitly instructing and supervising the robot. We evaluate \algabbr{} through user studies grounded in household manipulation tasks (e.g., assembling lunch bags) that measure the efficiency of the collaboration, as well as features such as perceived helpfulness, ease of use, and reliability. Our analysis suggests that both meta-prompting and proactivity are critical, resulting in 38.7\% faster task completion times and 31.9\% less user burden relative to non-active baselines.\footnote{Videos and Supplementary Material: \url{https://provox-2025.github.io}

Digital Object Identifier (DOI): see top of this page.}

\end{abstract}


\section{Introduction}
\label{sec:introduction}

\begin{figure}[t]
    \centering
    \includegraphics[width=\linewidth]{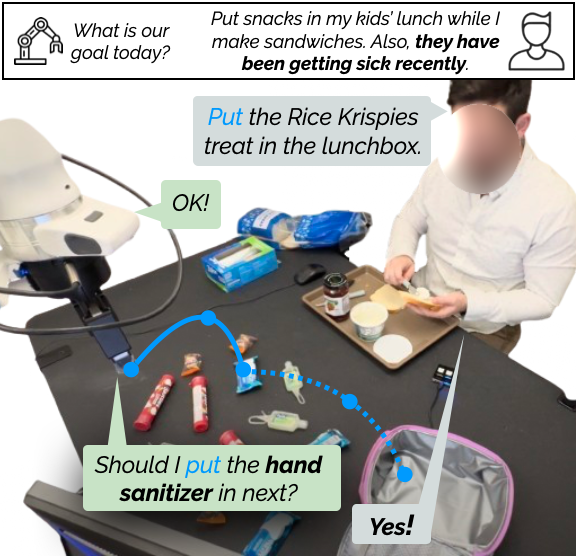}
    \vspace*{-6mm}
    \caption{We present \textbf{\algabbr{}} (``\algname{}''), a framework for personalization and proactive planning in the context of situated human-robot collaborations. In the first phase of a collaboration [\textbf{Top}], a human communicates their goals and distinct preferences, enabling the robot to \textit{personalize}. Throughout the rest of the collaboration [\textbf{Bottom}], the robot continues to incorporate and anticipate their partner's intent to \textit{proactively suggest helpful actions} (e.g. ``Should I put the hand sanitizer in next?'') ahead of explicit instruction, reducing the user's mental load while they assemble the sandwich.}
    \label{fig:front-fig}
    \vspace*{-7mm}
\end{figure}

\begin{figure*}[t]
    \centering
    \includegraphics[width=\linewidth]{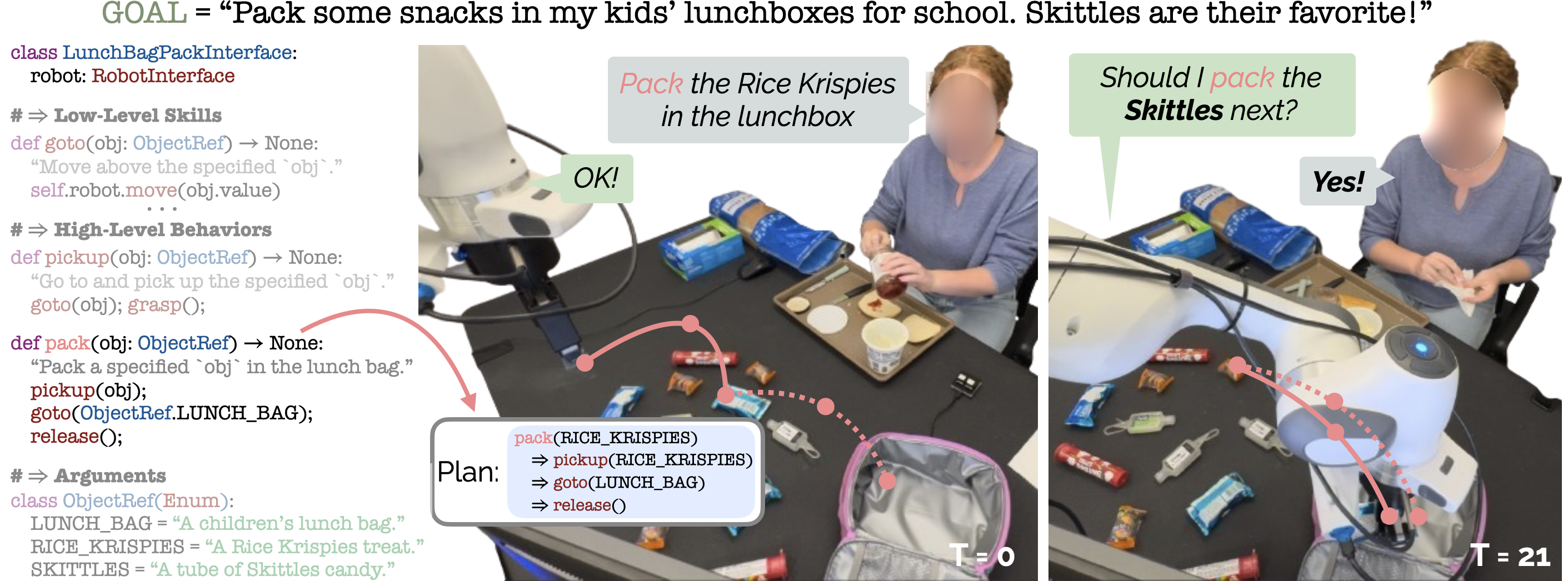}
    \vspace*{-5mm}
    \caption{\textbf{\algabbr{} Motivating Example.} Existing frameworks for situated human-robot collaboration tend to assume static, hard-coded APIs to inform task planning ({\color{SkeletonGray} gray}) that cannot be adapted to new individuals with distinct objectives and preferences. Instead, \algabbr{} allows users to provide high-level goals [\textbf{Top}] and define task-relevant actions (e.g., \texttt{pack} on the [\textbf{Left}]). This enables \textit{personalization} to a user's specific vocabulary and commands [\textbf{Middle}], and \textit{proactive planning} [\textbf{Right}], where the robot suggests helpful behaviors to accomplish the goal.}
    \label{fig:motivating-example}
    \vspace*{-6mm}
\end{figure*}


Collaborative robots must be able to continually infer their partner's intent, adapting from this information to personalize and proactively suggest helpful actions. This is especially true in the context of \textit{situated human-robot collaboration} \citep{hoffman2004collaboration, ajoudani2017collaboration, brawer2018situated, shaikewitz2023bite}, where robots and humans share the same physical space -- a setting that spans increasingly important applications such as household robotics, elderly or assistive care, warehouse manufacturing, and robot-assisted surgery, amongst others \citep{kruger2009cooperation, sharkawy2021survey}. Across these applications, effective collaboration is challenging due to the sheer diversity of human partners, each with their own distinct goals and preferences.

Consider the example in \autoref{fig:front-fig} of a household robot working with a person to assemble multiple lunch bags on a busy morning. Different people express different constraints on the overarching task -- for example, the person in \autoref{fig:front-fig} wants each bag to contain snacks, a sandwich, and a hand sanitizer because his kids have been sick, while in \autoref{fig:motivating-example}, another person needs the bag to contain Skittles, their favorite snack. In both cases, the human needs to perform the dexterous, fine-grained task of making the sandwich (i.e., grabbing two slices of bread, spreading the jelly and cream cheese, slicing off the crusts); however, without any other information, the division of work between the robot and human for the rest of the task is ambiguous. In such a scenario, a passive robot \citep{saycan2022google, wang2024mosaic, grannen2024vocal} might wait for explicit instructions, expecting the human to context-switch between making the sandwich and monitoring the robot's progress -- for example, repeatedly instructing the robot to ``put the Rice Krispies treat in the lunchbox'' followed by similar instructions for the hand sanitizer and candy (and again for the next lunch bag) -- a process that is as inefficient as it is frustrating. Instead, a more productive collaboration might start with a \textit{handshake}: an explicit protocol where the human iterates with the robot to build up a shared understanding of their intent. Doing so allows the robot to \textit{personalize} and adapt to each individual; now, as soon as the interaction starts, the robot might work from this shared understanding to \textit{proactively suggest a helpful plan}. From \autoref{fig:front-fig}, this might be as simple as extrapolating from the user's comment about his sick kids to suggest hand sanitizer as a possible addition to each lunch bag (``Should I put the hand sanitizer in next?'').


In this work, we formalize this process by introducing \textbf{\algabbr{}} (``\algname{}''), a new framework for developing personalizable and proactive collaborative robots that adapt online from language-based interactions with a partner. \algabbr{} builds on top of prior work for situated collaboration \citep{wang2024mosaic, grannen2024vocal} that leverage the commonsense priors and steerability of large language models \citep[LMs;][]{openai2023gpt4, saycan2022google, codeaspolicies, innermonologue} for task planning. Our first contribution is a  \textit{meta-prompting protocol} that equips users with a natural interface for not only communicating their overall objectives to the robot, but also for specifying concrete examples to seed the robot with an understanding of the user's distinct vocabulary and preferences (\autoref{fig:front-fig}; Top). We use the resulting prompt to condition a \textit{proactive language model planner} that anticipates what the robot should do next from the interaction context, suggesting helpful plans that work to alleviate user burden and improve the efficiency of the collaboration (\autoref{fig:front-fig}; Bottom).

We evaluate our technical contributions through two user studies. In our first study, we evaluate our meta-prompting protocol, performing a survey ($N = 26$) that demonstrates the flexibility and effectiveness of our proposed protocol relative to existing meta-prompting approaches. We find that our meta-prompting procedure universally improves our language model's ability to proactively suggest helpful plans while handling cross-user diversity relative to baselines. Finally, we perform a real-world within-subjects user study comparing \algabbr{} to a state-of-the-art passive, user-agnostic system \citep{grannen2024vocal} grounded in the lunch bag packing scenario seen in \autoref{fig:front-fig} and \autoref{fig:motivating-example}. We find that \algabbr{} enables 38.7\% faster collaborative task performance, with participants strongly preferring our system due to its ease of use (+27.3\%) and helpfulness (+18.4\%), as well as their willingness to use it again (+26.5\%).

\section{Related Work}
\label{sec:related-work}

\algabbr{} builds on prior work that propose new learning frameworks for situated human-robot collaboration, methods that leverage the commonsense reasoning and in-context learning ability of large language models for task planning, and general approaches for developing personalized and proactive robots in the context of human-robot interaction.

\smallskip

\noindent \textbf{Learning Frameworks for Situated Collaboration.} An expansive body of work frames human-robot collaboration as turn-taking, where a robot executes actions conditioned on a partner's prompt, spanning modalities such as natural language instructions \citep{tellex2020robonlp, brohan2023rt2} or gestures \citep{matuszek2014gesture, lin2023giraf}. Realizing the lack of adaptivity in these approaches, subsequent work develop methods for adapting robot behavior online, from more dynamic inputs such as real-time language corrections \citep{cui2023corrections, shi2024yell} or physical interventions \citep{bajcsy2017learning}. More recently, work such as MOSAIC \citep{wang2024mosaic} and Vocal Sandbox \citep{grannen2024vocal} extend such methods to develop fully-fledged systems that connect diverse modalities for interaction and teaching to build collaborative robots that can sustain long-horizon interaction for predefined tasks (e.g., cooking a fixed recipe). We build \algabbr{} on top of these works, specifically extending the Vocal Sandbox \citep{grannen2024vocal} framework to generalize to users with distinct goals and preferences while also enabling proactive planning (\autoref{fig:motivating-example}). 

\smallskip

\noindent \textbf{Language Models as Task Planners.} 
Many approaches that map language to robot behavior do so by leveraging the commonsense priors and steerability afforded by large, pretrained language models (LMs). These approaches often use LMs for \textit{task planning}, mapping complex user instructions to structured intermediate representations \citep{saycan2022google, singh2022progprompt} that are used to inform robot behavior. 
Many recent methods formalize task plans as executable programs \citep{codeaspolicies, liu2024okrobot, varley2024embodied}, using LMs to generate sequences of function calls subject to a predefined API -- for example, a Python class defining primitive robot behaviors such as \texttt{grasp()} or \texttt{pickup(obj: ObjectRef)}. 

While specifying task plans as executable programs offers benefits such as interpretability, runtime validation (i.e., always ensuring that generated task plans compile), and extensibility \citep{groundeddecoding, arenas2024cap, openai2023functioncalling}, code-based planners introduce new challenges when used in the context of human-robot collaboration. Specifically, a new partner trying to instruct such a robot needs a robust understanding of the robot's underlying API \textit{as well as} a rudimentary understanding of what language instructions invoke different functionality. Prior work attempts to overcome these issues through solutions that either make exclusionary assumptions about the types of people who can use such systems \citep[e.g., assuming enough code and LM literacy to understand what to say to induce a given behavior;][]{codeaspolicies}, or otherwise place undue burden on the user to work out what they can or cannot say through trial-and-error, or through expensive ``practice'' sessions with the robot \citep{wang2024mosaic, grannen2024vocal}. This need to ``naturalize'' \citep{wang2017naturalizing} a planning interface to a given person motivates the meta-prompting contributions in this work.

\smallskip

\noindent \textbf{Personalization and Proactivity in HRI.} The proactive planning component of \algabbr{} is informed by a wealth of work that learns to infer a user's intent through interaction. For example, algorithms for preference-based learning \citep{jain2015learning, biyik2018batch, biyik2021aprel} model the space of user intents by parameterizing reward functions based on a predefined set of features, resolving ambiguity by actively querying the user (e.g., asking them to rank or score different robot behaviors), often while minimizing some auxiliary cost (e.g., frequency of user queries, time to recover the ground-truth reward, etc.). Other work seeks to explicitly learn predictive models of user behavior \citep{nikolaidis2014efficient, sadigh2016information, javdani2018shared} from a combination of offline and online interaction data; by predicting what the user might do next, robots can proactively choose their actions in a way that offers maximal assistance, while minimizing any associated cost. While we do not explicitly learn models of user behavior or reward in this work, our proactive planner tries to infer actions based on the full interaction history -- a history that encompasses the full sequence of actions taken by both the human and the robot.

\section{Problem Formulation}
\label{sec:problem-formulation}

\begin{figure*}[t]
    \centering
    \includegraphics[width=\linewidth]{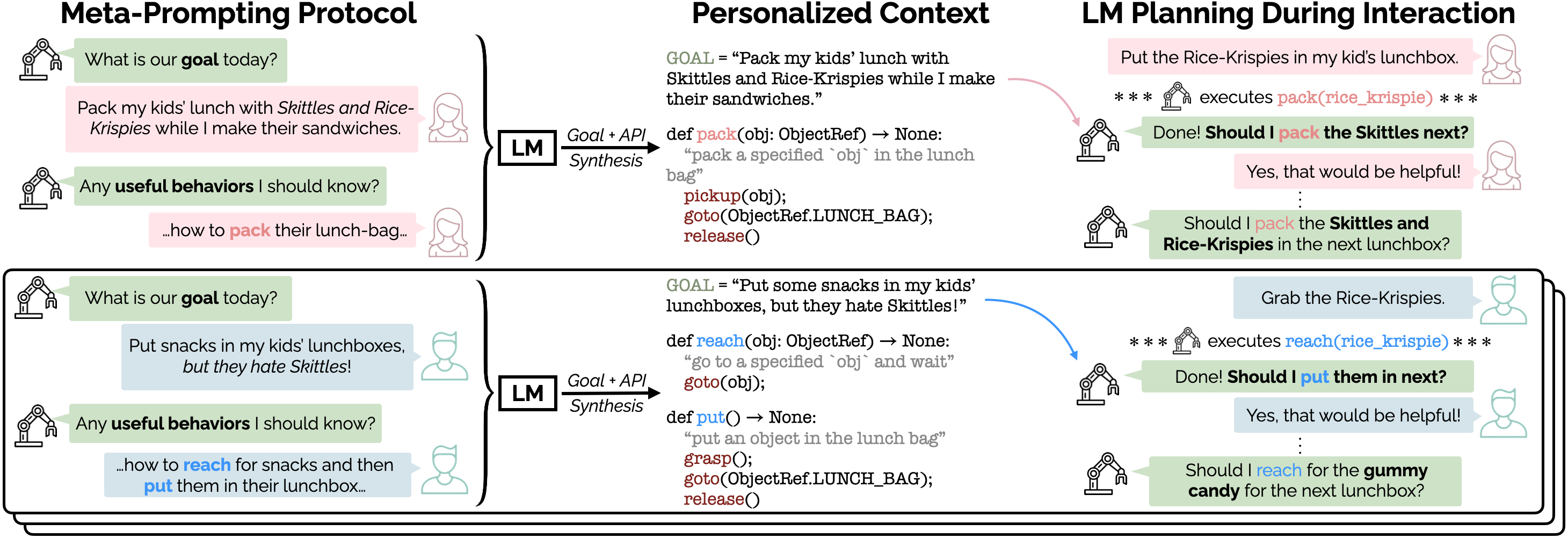}
    \vspace*{-5mm}
    \caption{\textbf{Meta-Prompting Protocol \& Proactive Planning.} \algabbr{} develops a novel meta-prompting protocol to collect two critical pieces of information from an individual: their specific goal, and an API of useful behaviors. Crucially, each user has a distinct set of preferences, yielding different goals and behaviors. The female user [\textbf{Top-Left}] wants her children's lunch to contain Skittles and Rice-Krispies, and teaches the robot to \texttt{pack} objects, with full confidence in its ability to identify, grasp, and move objects. In contrast, the male user [\textbf{Bottom-Left}] is more hesitant in trusting the robot; thus, he separates pick-and-place into two parts: a motion to \texttt{reach} for an object (moving above it without grasping), followed by a \texttt{put} behavior to complete the motion. The language model task planner then leverages this meta-prompted context to proactively suggest helpful, personalized behaviors over the course of the interaction [\textbf{Right}].}
    \label{fig:meta-prompting}
    \vspace*{-6mm}
\end{figure*}

\algabbr{} builds on Vocal Sandbox \citep{grannen2024vocal}, a framework for situated collaboration that allows human partners to teach robots new behaviors and concepts \textit{online}, during an interaction. We first formalize and highlight the key mechanisms of Vocal Sandbox, while the following section (\autoref{sec:proactive-voice}) introduces our novel contributions for personalization and proactive planning.

\subsection{Vocal Sandbox -- Preliminaries}
\label{subsec:vs-preliminaries}

Vocal Sandbox consists of a high-level language model (LM) planner and a family of low-level skills. The LM task planner maps spoken utterances from the human collaborator to code that subscribes to a predefined API; the Python-based API we use for our main user study is shown in \autoref{fig:motivating-example} (Left). Each function in the API (e.g., \texttt{goto(obj: ObjectRef)}) corresponds to an individual skill that is parameterized by arguments as shown in the function signature. ``Executing'' a function means rolling out the corresponding skill with the provided arguments (e.g., \texttt{goto(LUNCH\_BAG)}). 

A critical affordance uniquely provided by Vocal Sandbox is the ability for human collaborators to use language to teach the robot new behaviors (i.e., functions) online, throughout the course of an interaction. \autoref{fig:motivating-example} provides a concrete example in which the robot's base API ({\color{SkeletonGray} gray}) consists of simple motion primitives such as \texttt{goto(obj: ObjectRef)}, and \texttt{pickup(obj: ObjectRef)}, as well as operations for opening/closing the gripper. While this base API covers the space of motions the robot can perform, it is not ergonomic or natural -- a limitation demonstrated in \autoref{fig:motivating-example} (Middle), when the user asks the robot to ``pack the Rice Krispies in the lunchbox.'' To teach this new behavior, the user verbally \textit{decomposes} their utterance into a sequence of functions that already exist in the API -- in this case, via the program \texttt{pickup(RICE\_KRISPIES); goto(LUNCH\_BAG); release()}. Critically, after providing this decomposition, the Vocal Sandbox task planner uses the initial utterance and the corresponding program to \textit{synthesize} a new function \texttt{pack(obj: ObjectRef)} -- with the appropriate type signature and documentation -- adding it to the API so that it can immediately be used for the remainder of the interaction.

In \citet{grannen2024vocal}, users can teach new behaviors during a collaboration. However, this procedure requires substantial user effort during an interaction -- users must continually instruct the robot by continuously giving commands and decomposing unknown behaviors. Instead, \algabbr{} proposes a meta-prompting protocol, where users provide context and preferences to naturalize the system prior to task execution. Then, a proactive LM planner consumes this context to suggest helpful actions during a collaboration.


\subsection{Formalizing Task Planning and Teaching}
\label{subsec:vs-formalism}

Vocal Sandbox formalizes task planning as conditional language generation; 
here, we define notation for this problem setting. In \autoref{sec:proactive-voice}, we discuss \algabbr{} also using this notation.

At a given interaction step $t$, the language model $\text{LM}_\theta$ attempts to generate a programmatic plan $p_t$ conditioned on the user's natural language utterance $u_t$, the full interaction history $h_t = [(u_1, p_1), (u_2, p_2), \ldots (u_{t-1}, p_{t-1})]$, the current API $\Lambda_t$, and a global collaboration goal prompt $u_\text{fixed}$. Critically, $u_\text{fixed}$ is hand-designed by \citet{grannen2024vocal} and \textit{\textbf{held fixed for all users}}, serving to outline the full scope of the collaboration. This includes the specific task the robot and user are to complete, examples of possible language instructions, and expectations of how the robot should behave. Establishing a meta-prompting protocol for personalizing the goal prompt $u_\text{prompt}^k$ to a specific individual $k$ with distinct preferences and goals is a key contribution of \algabbr{} (\autoref{subsec:meta-prompting}). 

Teaching is formalized as a separate conditional generation task. Given an example $(\hat{u}_t, \hat{p}_t)$ consisting of a trigger utterance $\hat{u}_t$ (e.g., ``Pack the Rice Krispies in the lunchbox'' from \autoref{subsec:vs-preliminaries}), and the corresponding program decomposition $\hat{p}_t$ (e.g., \texttt{pickup(RICE\_KRISPIES); goto(LUNCH\_BAG); release()}), the goal is to \textit{synthesize} a new function to be added to the robot's API $\Lambda_{t+1}$. As described in \autoref{subsec:vs-preliminaries}, the output of the synthesis step is a new function name (e.g., \texttt{pack}), type signature (e.g., \texttt{(obj: ObjectRef) $\rightarrow$ \texttt{None}}), informative documentation string (e.g., \textit{``Pack a specified object in the lunch bag''}), and function body (e.g., \texttt{pickup(obj); goto(LUNCH\_BAG); release()}). To generate the function type signature and body from program decomposition $\hat{p}_t$, Vocal Sandbox employs prior unification-based algorithms for program induction \citep{wang2017naturalizing, karamcheti2020decomposition}, while the language model $\text{LM}_\theta$ generates both the function name and documentation from the interaction 
-- see \citet{grannen2024vocal} for further detail.

\algabbr{} employs a similar teaching approach to Vocal Sandbox with one key difference: individual users can teach behaviors \emph{before} a physical interaction begins. Teaching in advance of an interaction yields a more capable robot partner at task time, maximizing efficiency during costly robot interactions. We describe \algabbr{}'s novel contributions further in \autoref{sec:proactive-voice}.

\section{\algabbr{} -- Personalization \& Proactivity}
\label{sec:proactive-voice}

To enable personalization and proactive planning, \algabbr{} introduces two novel contributions: 1) a \textit{meta-prompting protocol} (\autoref{subsec:meta-prompting}) that enables individual users to communicate their distinct objectives, preferences, and expectations to the robot, and 2) a \textit{proactive language model planner} that suggests helpful actions ahead of explicit instructions (\autoref{subsec:proactive-planning}).

\subsection{Designing a Meta-Prompting Protocol}
\label{subsec:meta-prompting}

As discussed in \autoref{subsec:vs-formalism}, a key limitation of Vocal Sandbox and similar systems \citep{grannen2024vocal, wang2024mosaic, liu2024okrobot} is the reliance on a hand-designed global prompt $u_\text{fixed}$ and base API $\Lambda_\text{base}$ that outlines the full scope of a collaboration -- a scope that encompasses the specific task to complete, examples of ``valid'' language instructions, and expected robot behaviors. Beyond the question of generalizing to different users with distinct goals and preferences, relying on this global prompt also \textit{unfairly homogenizes the nature of a human-robot collaboration}; different users are expected to use the same vocabulary (e.g., behavior or object names), adopt the same roles, and use the same conventions prescribed by the system designers, which can be damaging in the presence of diverse users who want to assert different amounts of autonomy or trust. \autoref{fig:meta-prompting} shows an example of these disparities. Not only do the users have unique goals and preferences as to what should go in each lunch box, but they even define different behaviors for placing objects in each bag; rather than let the robot perform the entire pick-and-place motion continuously, the second user wants to verify how individual objects will be grasped prior to transferring them to the lunch box (\autoref{fig:meta-prompting}; Bottom).

We address these questions of per-user personalization by defining a novel \textit{meta-prompting protocol} that allows an individual $k$ to naturally communicate their preferences and goals to arrive at a personalized prompt $u^k_\text{prompt}$ and API $\Lambda^k$. We express this protocol via a graphical user interface, with an overview of the core features visualized in \autoref{fig:meta-prompting}.
\footnote{We provide additional figures and code on \href{https://provox-2025.github.io}{our project page}.} 
Intuitively, the interface gives users the ability to directly work with the language model task planner, iteratively building $u^k_\text{prompt}$ and $\Lambda^k$ through a ``query-driven development'' workflow \citep{beck2002tdd, liu2021prompting}. 

A user starts by describing their individual goal and preferences in natural language (e.g., ``Pack my kids' lunch with Skittles and Rice-Krispies while I make their sandwiches'' in the \texttt{GOAL} field of \autoref{fig:meta-prompting}; Middle), which the LM planner interprets to set $u^k_\text{prompt}$ for the personalized context. Individuals then \textit{iteratively test and verify the outputs of the LM planner} (in isolation, prior to interacting with the physical robot) by providing test utterances such as ``can you put the cereal bar in the bag?'' or ``how about packing the hand sanitizer?''. Before teaching any new behaviors, the LM may output a valid plan given its reasoning and common sense abilities (i.e., \texttt{pickup(cereal)}, \texttt{goto(lunchbox)}, \texttt{release()}). However, in most instances, the LM fails to initially generate a satisfying plan. Users then have the option to explicitly teach new API functions themselves by specifying: (1) function name (e.g., ``pack''), (2) expected behavior (e.g., ``packing food for lunch''), and (3) the function body through a drop-down menu (e.g., \texttt{pickup(food)}, \texttt{goto(lunchbox)}, \texttt{release()}). Users may choose to iterate on their taught behaviors, trying out what they have taught to see if the LM planner matches the expected behavior (e.g., ``Put the cereal bar in my lunch.'' maps to \texttt{pack(cereal)}). Giving users explicit insight into the API in this manner is a key difference from the LM-driven function synthesis procedure in \citet{grannen2024vocal}; we evaluate the benefits of doing this with user studies (\autoref{subsec:study-meta-prompting}). 

We provide users with the ability to view the history of taught actions and LM outputs, as well as freely edit and delete their goal text and taught functions. For simplicity, we do not track the entire interaction history $h_t = [(u_1, p_1), (u_2, p_2), \ldots (u_{t-1}, p_{t-1})]$ when using the LM to generate plans. Note that teaching is not limited to definitions via this interface; if the user realizes they want to define new behaviors while physically interacting with the robot, they can do so via the teaching-via-synthesis procedure in \autoref{subsec:vs-formalism}. 

We provide the context output (a goal $u^k_\text{prompt}$ and an API $\Lambda^k$) to the LM planner, thereby enabling it to proactively suggest actions (see \autoref{fig:meta-prompting}; Right). 
We emphasize the novelty of our meta-prompting protocol for collaborative task planning: where prior situated collaboration works lack personalization~\cite{grannen2024vocal, wang2024mosaic}, \algabbr{}'s meta-prompting protocol enables users to communicate their personal objectives and preferences \textit{before an interaction begins}.
We evaluate the impact of meta-prompting relative to global prompting via an isolated, component-wise user study (\autoref{subsec:study-meta-prompting}), and the impact of meta-prompting in the context of a full system user study (\autoref{subsec:study-provox}).

\subsection{Proactive Planning}
\label{subsec:proactive-planning}

Another limitation of prior systems such as Vocal Sandbox is the lack of an ability to proactively suggest plans, preventing the robot from assuming more autonomy and initiative over the course of a collaboration. The robot executes the user's instruction and then waits idly for the next instruction (e.g. in \autoref{fig:meta-prompting}, ``Put the Rice-Krispies in my kids lunchbox.'' would invoke only \texttt{pickup(rice\_krispies)}). This limitation again stems from the expensive upfront cost of having new users build a mental model and ``naturalize'' \citep{wang2017naturalizing, nikolaidis2012human} to a system reflecting the conventions and expectations of another person (i.e., the initial system designer). The second contribution of \algabbr{} builds on top of the grounded understanding of an individual's goals, preferences, and desired behaviors obtained as a result of the meta-prompting procedure, yielding a \textit{proactive language model planner} -- a LM planner that pairs the commonsense abilities embedded in LMs with the information encoded in $u^k_\text{prompt}$ and $\Lambda^k$ to progressively suggest actions that help maximize the efficiency of the collaboration.

We implement proactivity as a straightforward extension of the base task planner described in \autoref{subsec:vs-formalism}. After each executed task plan $p_t$, we invoke the LM planner again by prompting it with the following trigger string: ``Propose an action to perform next to perform [user-provided goal $u^k_\text{prompt}$].''
For example, in \autoref{fig:meta-prompting} the robot proactively suggests helpful actions after completing each instruction -- from finishing the current lunchbox (``Should I pack the Skittles next?'') to proactively suggesting longer plans for packing a full lunchbox (``Should I pack the Skittles and the Rice Krispies in the next lunchbox?''). 
We do this continually at each interaction step, using the full interaction history (and potentially updated API) to shape the planner's suggested behaviors. For safety and transparency, we do not automatically execute proposed plans. Rather, the robot displays its intended plan on an interface and vocalizes it, then gates on user confirmation before execution. We note that if the proactive LM planner fails, users retain the ability to instruct the robot as in \citet{grannen2024vocal}.

The novelty of \algabbr{}'s proactive planner is that it actively suggests helpful actions that we show minimize robot downtime and streamline collaborations
through a full system user study (\autoref{subsec:study-provox}), evaluating \algabbr{} against a non-active Vocal Sandbox baseline. We measure the efficiency of the collaboration as well as qualitative metrics such as ease of use, predictability, and perceived helpfulness.

\section{Implementation \& Reproducibility}
\label{sec:implementation-reproducibility}

\begin{figure*}[t]
    \centering
    \includegraphics[width=\linewidth]{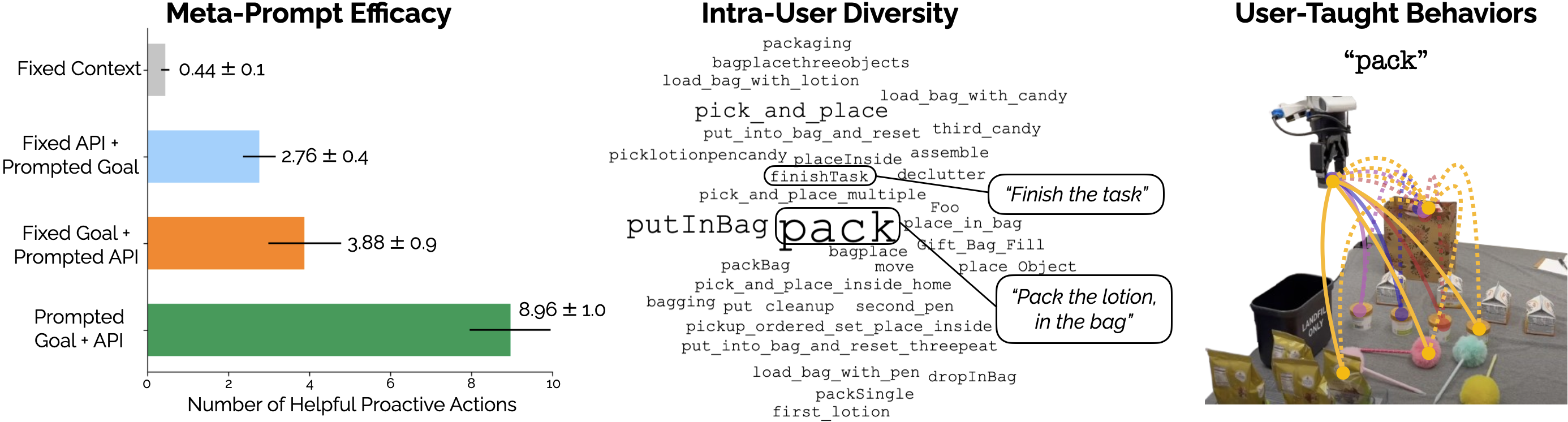}
    \vspace*{-5mm}
    \caption{\textbf{Meta-Prompting User Study Results.} 
    We present results evaluating the efficacy of \algabbr{}'s meta-prompt protocol through a $N=26$ user study, as well as visualize the diversity of user-defined behaviors. We report the mean and standard error values of the number of helpful, proactive actions suggested by the LM planner given full, partial, or no access to the meta-prompted context [\textbf{Left}]. We observe that both parts of the meta-prompted context (goal and API) are crucial for downstream proactive suggestion capabilities. 
    We also highlight the intra-user diversity with a word cloud of 33 user-taught behavior names as well as two corresponding commands [\textbf{Middle}], with the majority of the behavior names being unique with the exception of \texttt{pack} (5X), \texttt{putInBag} (3X), and \texttt{pick\_and\_place} (2X). Interestingly, while five behavior instances are named \texttt{pack}, each corresponds to a unique program decompositions, visualized in different colors on the [\textbf{Right}], underscoring the need for personalization -- even \texttt{pack} means different behaviors to different people!}
    \label{fig:meta-prompting-user-study}
    \vspace*{-6mm}
\end{figure*}

In this work, we implement the \algabbr{} framework mostly following the design decisions in \citet{grannen2024vocal} with minor changes. First, due to OpenAI's deprecation warning of the GPT-3.5 Turbo suite of language models used in the original Vocal Sandbox work \citep{grannen2024vocal}, we adopt the more recent GPT-4 Turbo \citep[v04-09-2024;][]{openai2023gpt4} as our base language model $\text{LM}_\theta$. As our full system user study (\autoref{subsec:study-provox}) focuses on a different application (lunch bag packing), we define a new base API to reflect the new objects and motion primitives; the full Python implementation can be found on our website. As in \citet{grannen2024vocal} we format the API as an object-oriented Python API, formatted as a Markdown code block. We use the function-calling capabilities provided by OpenAI to constrain the language model to generate valid programs.

\medskip

\noindent \textbf{Robot Platform \& Motion Primitives}. We use a Franka Panda fixed-arm manipulator equipped with a Robotiq 2F-85 gripper following the hardware specification in \citet{khazatsky2024droid}. We also assume an extrinsics-calibrated ZED 2 stereo camera to obtain point cloud observations. As participants and the robot share a physical workspace, we run a compliant controller with a contact force threshold of 40 N, and torque threshold of 30 Nm. We implement two software-based kill switches (one controlled by the participant, the other by the study proctor), as well as two hardware emergency stops. We implement our \texttt{goto(obj)} and \texttt{pickup(obj)} motion primitives by identifying heuristic offsets relative to the 3D centroid of each object in the robot's coordinate frame. We obtain these centroid coordinates from off-the-shelf vision models following \citet{grannen2024vocal}, first obtaining a 2D segmentation mask via FastSAM \citep{zhao2023fast}, then retrieving a point cloud via backprojection through our calibrated camera.

\section{User Studies}
\label{sec:user-studies}

We evaluate \algabbr{} through two user studies. The first study (\autoref{subsec:study-meta-prompting}; $N = 26$) serves to evaluate the meta-prompting protocol in isolation, evaluating the proposed procedure's ability to capture different user preferences and effectively inform proactive planning in a controlled experiment. We then perform a full-system study on a real-world robot platform (\autoref{subsec:study-provox}; $N = 9$) that has participants evaluate \algabbr{} against a (non-proactive) Vocal Sandbox system for a collaborative lunch bag packing application. All studies were IRB approved.

\subsection{Meta-Prompting for Grocery Bagging}
\label{subsec:study-meta-prompting}

In this study, participants evaluate our meta-prompting protocol by attempting to specify a goal and API to accomplish a given task (shown to participants as a video of a robot rolling out in a controlled environment). We aim to measure the diversity of goals and behaviors our procedure can cover, as well as quantitative metrics that speak to how well the resulting meta-prompted contexts inform proactive planning. 

\smallskip

\noindent \textbf{Participants and Procedure}. We conducted this study with a population of $N = 26$ participants (within-subjects -- 9 female, 17 male; ages between 21 and 69) with some amount of prior experience working with robots (an average self-reported experience score of $3.65 / 7$). After providing informed consent, participants were shown a 26 second video of a Franka Emika Panda robot bagging three items (a bottle of lotion, a pen, and candy) and tasked with reproducing the behavior in as few language instructions as possible. Participants were given access to a list of initial robot behaviors (e.g., \texttt{pickup}), a set of example ``trigger'' utterances (e.g., ``pick up the ...''), and the corresponding robot demonstration videos. We asked each participant to engage with the meta-prompting protocol over the course of 20 minutes, working to produce a goal $u_\text{prompt}^k$ and API $\Lambda^k$ they felt best represented the initial video.

\smallskip

\noindent \textbf{Independent Variables: Meta-Prompt Information}. We evaluate the efficacy of our meta-prompting protocol by using each participant's resulting prompt (e.g., the context $u_\text{prompt}^k$, $\Lambda^k$) to condition the proactive language model planner; if the participant was able to properly communicate their goal and expected behaviors, the planner should be able to suggest a maximal sequence of behaviors that make progress towards replicating the actions taken in the provided robot video. We compare our proposed protocol (\textit{Prompted Goal + API}) to three baselines that ablate specific components. \textit{Fixed Goal + Prompted API} conditions the language model planner with a fixed, neutral goal  $u_\text{fixed} = $ ``to help the user with a tabletop task,'' but using the participant's taught behaviors $\Lambda^k$. \textit{Fixed API + Prompted Goal} does the opposite, using the participant's prompted goal $u_\text{prompt}^k$, but does not allow access to the taught behaviors, instead using the base API $\Lambda_\text{base}$. Finally, \textit{Fixed Context} assumes a neutral goal and fixed API.

\smallskip

\noindent \textbf{Dependent Measures}. We measure the efficacy of each approach by generating the proactive task plan $p_\text{pro}$ following \autoref{subsec:proactive-planning}. We compare $p_\text{pro}$ with the ground truth task plan $p_\text{ref}$ depicted in the initial demonstration, reporting the count of overlapping function invocations (individual behaviors represented in the reference video) as \textit{Meta-Prompt Efficacy}. Large overlaps indicate that the proactive planner and the corresponding meta-prompting method are effective in suggesting helpful actions that mimic the ground truth task plan $p_\text{ref}$. 

\smallskip

\noindent \textbf{Hypotheses}. This intuition informs our first hypothesis (\textbf{H1}) that proactive planners with access to parts of the meta-prompted context (\textit{Fixed API + Prompted Goal} and \textit{Fixed Goal + Prompted API}) are more effective in suggesting helpful actions than a fixed context planner (\textit{Fixed Context}). Furthermore, our second hypothesis (\textbf{H2}) affirms that the proactive planner with full access to the meta-prompted context (\textit{Prompted Goal + API}) would be more effective in suggesting helpful actions than any other method.

\smallskip

\begin{figure*}[t]
    \centering
    \includegraphics[width=\linewidth]{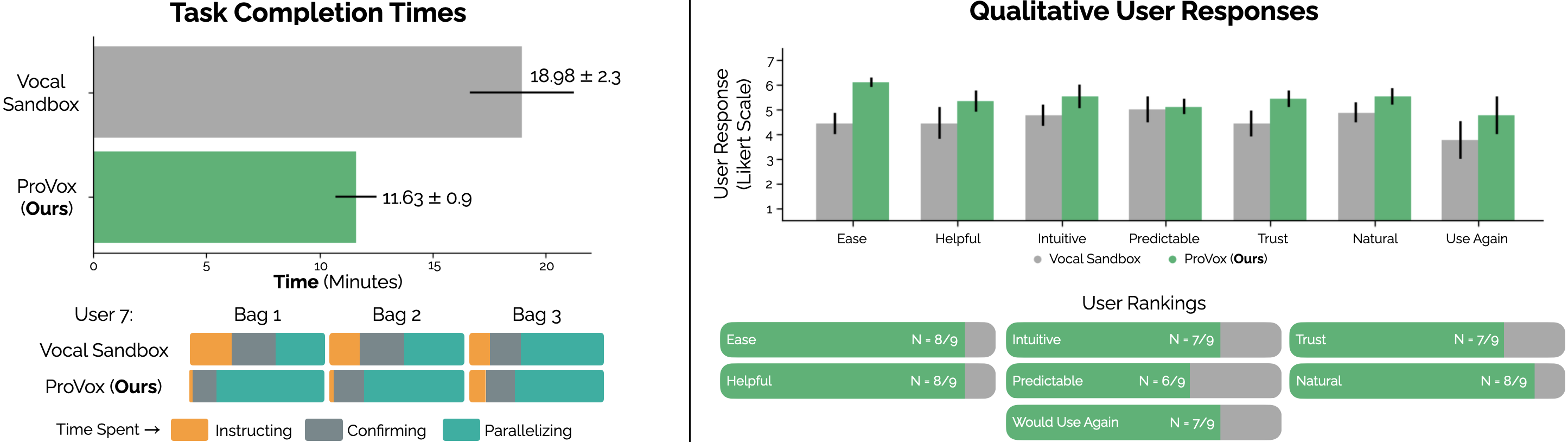}
    \vspace*{-5mm}    
    \caption{\textbf{Full System User Study Results.} We present quantitative and qualitative results from our real-world user study of collaborative lunch bag packing. We report the average and standard error task completion times for users collaborating with a \algabbr{} or a Vocal Sandbox system to pack three lunch bags [\textbf{Top, Right}]. We further visualize the breakdown of how a specific user spends their time throughout an interaction [\textbf{Bottom, Right}], highlighting the minimal instruction needed in the \algabbr{} collaboration. On the [\textbf{Top, Right}], we report the average and standard error user responses on a 1--7 Likert scale across seven questions. We also ask users to explicitly rank which of the two methods they favor across these seven categories [\textbf{Bottom, Right}]. We observe in both the responses that users strongly favor \algabbr{} over the baseline collaborator in terms of ease, helpfulness, trustworthiness, and willingness to use again.}
    \label{fig:physical-user-study}
    \vspace*{-6mm}
\end{figure*}

\noindent \textbf{Results}. We report \textit{Meta-Prompt Efficacy} for each of the methods in \autoref{fig:meta-prompting-user-study} (Left). Our graph clearly shows that the \algabbr{} meta-prompting protocol (\textit{Prompted Goal + API}) achieves the highest efficacy with an average of $8.96 \pm 1.0$ helpful actions proposed, supporting (\textbf{H2}). We also observe that the methods with partial access to the meta-prompted context, \textit{Fixed API + Prompted Goal} and \textit{Fixed Goal + Prompted API}, suggest an average of $3.88 \pm 0.9$ and $2.76 \pm 0.4$ helpful proactive actions respectively; this is higher than the average $0.44 \pm 0.1$ helpful actions suggested by the \textit{Fixed Context} method, supporting (\textbf{H1}). 

We run two-way repeated-measures ANOVA tests to assess the significance of these results. We find that having partial or full access to the meta-prompted context has a statistically significant effect ($p \leq 0.05$) on meta-prompt efficacy compared to using a fixed context. We find that using the full meta-prompted context has a significant effect ($p \leq 0.05$) on meta-prompt efficacy compared to partial access. 

Beyond efficacy, we visualize the diversity of taught behaviors across participants in \autoref{fig:meta-prompting-user-study}, highlighting the need for per-user personalization. Across $N = 26$ participants, our user-specific meta-prompted contexts consist of $39$ unique behaviors with $33$ distinct names (\autoref{fig:meta-prompting-user-study}; Middle). Even within taught behaviors of the same name, participants elect to teach vastly different program decompositions. For example, there are five instances of taught behaviors with the most common name, \texttt{pack}, however there are three unique program decompositions of these behaviors  (visualized in~\autoref{fig:meta-prompting-user-study}; Right). 


\subsection{Full-System Evaluation: Lunch Bag Packing}
\label{subsec:study-provox}

In this full-system study, participants collaborate with a \algabbr{} system to pack three lunch bags. Each bag must contain two snack items, a bottle of hand sanitizer, and one prepared sandwich. Participants may interact with the robot to pack items in each lunch bag while they focus on the dexterous task of making the sandwich (i.e., spreading the jam and cream cheese, cutting off the crusts, and placing it a Ziploc bag). 

\smallskip

\noindent \textbf{Participants and Procedure}. We conducted the full-system study with a population of $N = 9$ participants (within-subjects -- 4 female, 5 male; ages between 25 and 28), again with a some amount experience of working with robots (an average self-reported experience of $3 / 7$). Each participant performed the task with each of two methods (\algabbr{} and Vocal Sandbox), with a random ordering across users. After providing informed consent, we provided each participant with a sheet describing the robot's base API (consisting of 5 initial behaviors) and teaching interfaces. We also offered the opportunity to try a practice task (``clear the table'' -- unrelated to the lunch bag packing task). To systematically evaluate different preferences, we assigned each participant 3 items to pack in the lunch bag (in addition to the prepared sandwich), from the set consisting of hand sanitizer, Rice Krispies treat, Skittles, and gummy candy. If applicable, participants were asked to engage with the meta-prompting protocol to collect their personalized context ahead of interacting with the robot.

\smallskip

\noindent \textbf{Independent Variables: Personalization \& Proactivity}. We compare \algabbr{} against an instantiation of a (non-proactive) Vocal Sandbox system \citep{grannen2024vocal}; running this head-to-head comparison allows us to evaluate our technical contributions while controlling for the remaining system capabilities.

\smallskip

\noindent \textbf{Dependent Measures}. We consider both objective and subjective metrics to evaluate our framework. For each method, we report the time needed to pack three lunch bags (\textit{Task Completion Times}) as well as the number of user or robot-lead plan proposals throughout the task. After completing the task, participants complete a survey of 7-point Likert scale questions to assess the qualitative characteristics of the given method: \textit{Easy to Use}, \textit{Helpful}, \textit{Intuitive}, \textit{Predictable}, \textit{Trust}, \textit{Natural}, and \textit{Willingness to Use Again}. After interacting with both methods, we additionally asked participants to explicitly rank each method subject to the same criteria. 

\smallskip

\noindent \textbf{Hypotheses}. Our first hypothesis (\textbf{H1}) asserts that participants will complete the lunch bag packing task faster with the \algabbr{} system compared to the Vocal Sandbox system; intuitively, a faster task completion time validates the overall utility of our proposed contributions. Furthermore, we affirm that (\textbf{H2}) the \algabbr{} system reduces the burden on each user to provide explicit instructions by proposing helpful plans. Finally, we expect that (\textbf{H3}) participants qualitatively prefer \algabbr{} to Vocal Sandbox across all criteria due to the utility provided by personalization and proactivity.

\smallskip

\noindent \textbf{Results}. We report quantitative and qualitative results in \autoref{fig:physical-user-study}. In collaborating with the \algabbr{} system, participants complete the task in an average of $11.63 \pm 0.9$ minutes, significantly faster than the $18.98 \pm 2.3$ minutes needed for the Vocal Sandbox system, supporting (\textbf{H1}). We note that these times only include the (situated) human-robot collaboration, and exclude the time spent in the meta-prompting interface; if we look at meta-prompting times alone, we see an average of $5.58 \pm 0.9$ minutes. Note that time spent engaging with the meta-prompting protocol is a fixed cost at the beginning of each interaction that does not scale with the horizon of the rest of the task. We additionally probe the proactivity of \algabbr{} by computing the ratio of user-initiated behaviors to robot-initiated behaviors (\autoref{fig:physical-user-study}; Left). Notably, participants accede to robot-proposed plans 31.9\% of the time, strongly supporting (\textbf{H2}). In further visualizing the breakdown of how participants spend time during the course of an interaction, we observe that users also spend 19.0\% \textit{less time} explicitly instructing \algabbr{} relative to Vocal Sandbox.

\autoref{fig:physical-user-study} (Right) plots the responses to the subjective questions. Participants strongly prefer \algabbr{} in terms of ease of use, helpfulness, intuitiveness, trustworthiness, naturalness and willingness to use again, supporting (\textbf{H3}). We also report user rankings for each of these criteria to explicit measure the trade-off between the two methods, where the trends are clearer.

We assess the significance of these results through one-way repeated-measures ANOVA tests. We find that collaborating with the \algabbr{} system has a statistically significant effect on task completion times, as well as user ratings for ease and helpfulness. The remaining subjective results are not statistically significant (an artifact of the relative small participant pool). In general, participants see clear gains from the personalization and proactivity provided by \algabbr{}, both in terms of collaboration efficiency and subjective perception.

\section{Discussion}
\label{sec:discussion}

We present \algabbr{}, a novel framework for developing personalizable and proactive robots in the context of situated collaboration. \algabbr{} proposes two novel technical contributions: (1) developing a meta-prompting protocol to personalize to an individual user's objectives and expected robot behaviors as well as (2) a proactive language model task planner that suggests helpful actions ahead of explicit user instruction. Through these contributions, \algabbr{} demonstrates the ability to not only generalize to a broad population of participants with distinct preferences, but also improve the efficiency of a human-robot collaboration. In general, \algabbr{} systems produce more easy to use (+27.3\%), helpful (+18.4\%) and trustworthy (+22.5\%) robot collaborators, with 38.7\% faster collaborative task performance and 31.9\% less user burden compared to state-of-the-art baselines \citep{grannen2024vocal}. That said, \algabbr{} is only a start, with multiple avenues for future work.



\smallskip

\noindent \textbf{Towards Diverse Feedback Modalities}. While \algabbr{} provides a framework for personalizing a language-based goal and API to a user, it fails to consider other feedback modalities that implicitly encode preferences, such as kinesthetic demonstrations or gestures. In general, a key limitation of \algabbr{} is its reliance on an \textit{ungrounded} language model; this opens up an interesting avenue of future work for integrating vision-language models \citep{liu2023llavav15, openai2023gpt4v} or even vision-language-action models \citep{kim2024openvla} for human-robot collaboration.

\bibliographystyle{IEEEtranN}

{\footnotesize
\bibliography{x-references}}

\end{document}